\pdfoutput=1
\documentclass{article}
\usepackage{iclr2026_conference,times}

\usepackage{amsmath,amsfonts,bm}

\def\eqref#1{equation~\ref{#1}}

\def\1{\bm{1}}

\DeclareMathAlphabet{\mathsfit}{\encodingdefault}{\sfdefault}{m}{sl}
\SetMathAlphabet{\mathsfit}{bold}{\encodingdefault}{\sfdefault}{bx}{n}

\usepackage{hyperref}
\usepackage{url}
\usepackage{graphicx}
\usepackage{booktabs}
\usepackage{algorithm}
\usepackage{algorithmic}
\usepackage{amsmath}
\usepackage{amssymb}
\usepackage{multirow}

\newcommand{\agenttrace}{\textsc{AgentTrace}}

\title{AgentTrace: Causal Graph Tracing for Root Cause Analysis in Deployed Multi-Agent Systems}

\iclrfinalcopy

\author{Zhaohui Geoffrey Wang \\
USC Viterbi School of Engineering \\
\texttt{zwang000@usc.edu} \\
ORCID: \href{https://orcid.org/0009-0006-1187-1903}{0009-0006-1187-1903}
}

\begin{document}

\maketitle

\begin{abstract}
As multi-agent AI systems are increasingly deployed in real-world settings---from automated customer support to DevOps remediation---failures become harder to diagnose due to cascading effects, hidden dependencies, and long execution traces. We present \agenttrace{}, a lightweight causal tracing framework for post-hoc failure diagnosis in deployed multi-agent workflows. \agenttrace{} reconstructs causal graphs from execution logs, traces backward from error manifestations, and ranks candidate root causes using interpretable structural and positional signals---without requiring LLM inference at debugging time. Across a diverse benchmark of multi-agent failure scenarios designed to reflect common deployment patterns, \agenttrace{} localizes root causes with high accuracy and sub-second latency, significantly outperforming both heuristic and LLM-based baselines. Our results suggest that causal tracing provides a practical foundation for improving the reliability and trustworthiness of agentic systems in the wild.
\end{abstract}

\section{Introduction}
\label{sec:intro}

Multi-agent systems powered by large language models (LLMs) have emerged as a powerful paradigm for solving complex tasks through agent collaboration~\citep{wu2023autogen, hong2024metagpt}. In these systems, specialized agents---such as planners, coders, reviewers, and executors---coordinate through message passing to accomplish goals that would be difficult for a single agent.

In deployed agent systems---such as automated customer support, DevOps remediation, or research assistants---failures often surface far downstream from their root causes. By the time an error is observed, multiple agents may have already acted on corrupted assumptions, making manual debugging slow and unreliable. The distributed and emergent nature of these workflows means that traditional debugging approaches, which examine individual components in isolation, fail to capture the cross-agent causal dependencies that lead to system-level failures.

We propose \agenttrace{}, a lightweight framework that addresses this challenge through three key contributions:

\begin{enumerate}
    \item \textbf{Causal Graph Construction}: We model multi-agent execution as a directed graph where nodes represent agent actions and edges capture information flow and causal dependencies.

    \item \textbf{Backward Tracing Algorithm}: Starting from the point of error manifestation, we trace backward through the causal graph to identify all potentially relevant upstream decisions.

    \item \textbf{Empirical Study of Failure Localization}: We demonstrate that lightweight causal tracing with interpretable positional and structural features can substantially improve debugging accuracy and latency in settings that resemble real-world agent deployments---without requiring expensive LLM inference at debugging time.
\end{enumerate}

\begin{figure}[t]
\centering
\includegraphics[width=\textwidth]{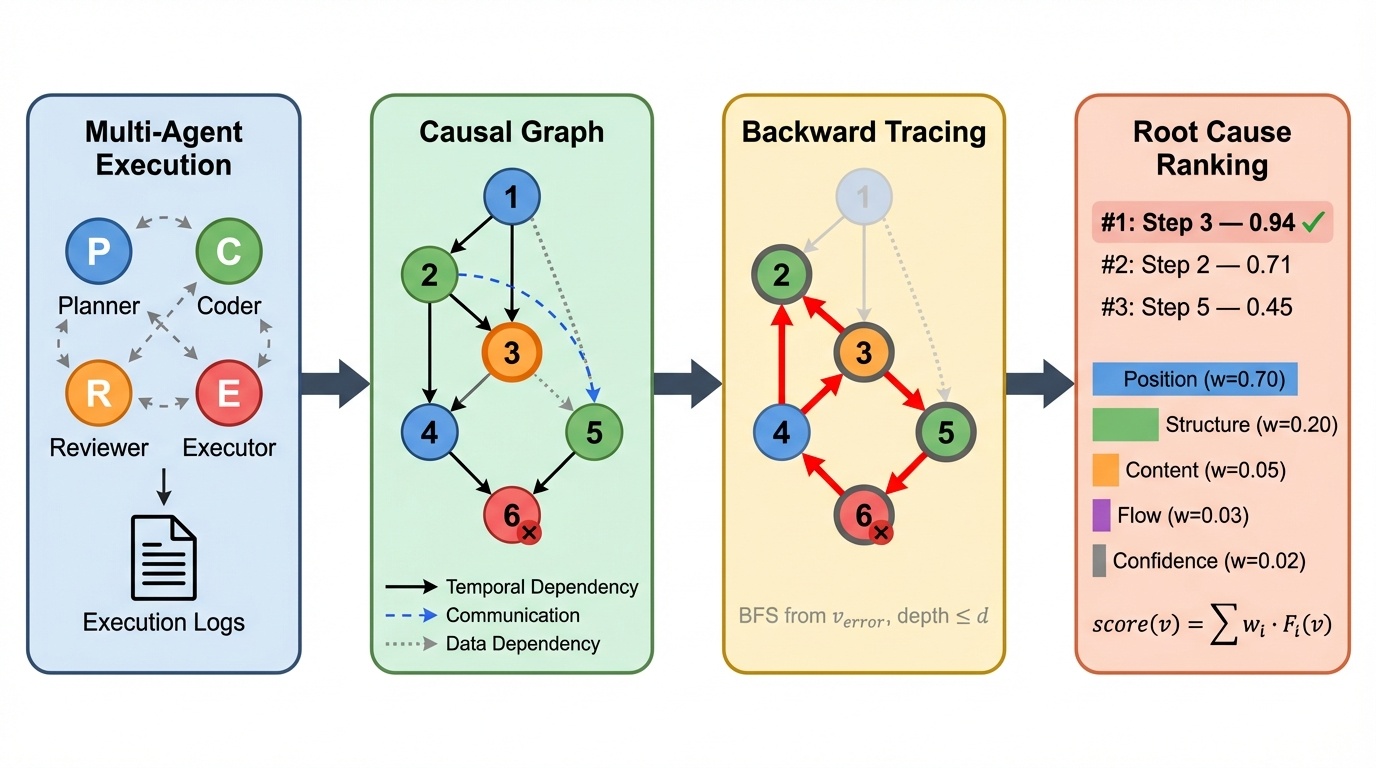}
\caption{Overview of the \agenttrace{} framework. (a)~Execution logs from a multi-agent workflow are collected. (b)~A causal graph is constructed with sequential, communication, and data dependency edges. (c)~Backward BFS from the error node identifies candidate root causes. (d)~Candidates are ranked using a weighted combination of five interpretable feature groups, with position features dominating ($w_p{=}0.70$).}
\label{fig:framework}
\end{figure}

While our benchmark scenarios are synthetically constructed, they are designed to capture common failure patterns observed in deployed multi-agent systems, such as early planning errors cascading through execution. Our evaluation across 550 failure scenarios in 10 domains shows that \agenttrace{} achieves high localization accuracy with sub-second latency, significantly outperforming both heuristic and LLM-based baselines. Figure~\ref{fig:framework} provides an overview of the framework.

\section{Related Work}
\label{sec:related}

\paragraph{Multi-Agent Systems.}
Recent frameworks like AutoGen~\citep{wu2023autogen} and MetaGPT~\citep{hong2024metagpt} enable sophisticated multi-agent collaboration, often building on reasoning-and-acting paradigms~\citep{yao2022react}. However, debugging support in these systems remains limited, typically relying on manual log inspection.

\paragraph{AI System Debugging.}
Prior work on AI interpretability has focused on neural network internals~\citep{simonyan2014deep, kim2018interpretability} or reasoning chains~\citep{wei2022chain}. Self-debugging approaches~\citep{chen2024selfdebugging} use LLMs to identify errors but require expensive inference calls and struggle with cross-agent issues.

\paragraph{Distributed Tracing.}
Systems like Jaeger~\citep{uber2017jaeger} and Zipkin~\citep{twitter2012zipkin} provide distributed tracing for microservices. We adapt these concepts for multi-agent LLM systems, where ``messages'' between agents carry semantic content rather than just request metadata.

\paragraph{Root Cause Analysis.}
Traditional RCA approaches use statistical methods~\citep{soldani2022anomaly} or graph algorithms~\citep{page1999pagerank}. Recent work explores LLM-based agents for root cause analysis in cloud systems~\citep{roy2024exploring}, but these methods are not designed for the specific structure of multi-agent workflows.

\section{AgentTrace Framework}
\label{sec:method}

\subsection{Problem Definition}

Given a multi-agent execution trace $T$ that terminates in an error state, our goal is to identify the \emph{root cause node}---the earliest decision point whose correction would prevent the error. Formally, let $G = (V, E)$ be a directed acyclic graph where:
\begin{itemize}
    \item $V = \{v_1, v_2, \ldots, v_n\}$ represents agent actions (tool calls, messages, decisions)
    \item $E \subseteq V \times V$ represents causal dependencies between actions
    \item $v_{\text{error}} \in V$ is the node where the error manifests
    \item $v_{\text{root}} \in V$ is the ground-truth root cause
\end{itemize}

\subsection{Causal Graph Construction}

We construct the causal graph from execution logs by identifying three types of edges:

\textbf{Sequential edges} connect consecutive actions by the same agent, capturing the agent's reasoning flow.

\textbf{Communication edges} connect message-send events to message-receive events between different agents.

\textbf{Data dependency edges} connect actions that produce data to actions that consume that data, identified through variable reference tracking.

\subsection{Backward Tracing}

Given an error node $v_{\text{error}}$, we perform breadth-first backward traversal to collect all ancestor nodes within a specified depth limit:

\begin{algorithm}[h]
\caption{Backward Tracing}
\begin{algorithmic}[1]
\REQUIRE Error node $v_{\text{error}}$, graph $G$, max depth $d$
\ENSURE Candidate set $C$
\STATE $C \leftarrow \{v_{\text{error}}\}$
\STATE $\text{frontier} \leftarrow \{v_{\text{error}}\}$
\FOR{$i = 1$ to $d$}
    \STATE $\text{new\_frontier} \leftarrow \emptyset$
    \FOR{$v \in \text{frontier}$}
        \FOR{$u \in \text{parents}(v)$}
            \IF{$u \notin C$}
                \STATE $C \leftarrow C \cup \{u\}$
                \STATE $\text{new\_frontier} \leftarrow \text{new\_frontier} \cup \{u\}$
            \ENDIF
        \ENDFOR
    \ENDFOR
    \STATE $\text{frontier} \leftarrow \text{new\_frontier}$
\ENDFOR
\RETURN $C$
\end{algorithmic}
\end{algorithm}

\subsection{Node Ranking Algorithm}
\label{sec:ranking}

We rank candidate nodes using a weighted linear combination of five feature groups:

\begin{equation}
\text{score}(v) = \sum_{i \in \{p, s, c, f, e\}} w_i \cdot F_i(v)
\end{equation}

Each group score $F_i(v)$ is the mean of its constituent normalized features:
\begin{equation}
F_i(v) = \frac{1}{|G_i|} \sum_{j \in G_i} \hat{f}_j(v), \quad \hat{f}_j(v) = \frac{f_j(v) - \min_{u} f_j(u)}{\max_{u} f_j(u) - \min_{u} f_j(u) + \epsilon}
\end{equation}

where $\epsilon = 10^{-8}$ prevents division by zero. The weights $w_i$ are determined via grid search on a held-out validation set of 50 scenarios (see Appendix~\ref{app:features}). Table~\ref{tab:feature_list} lists all 17 features across the five groups.

\begin{table}[t]
\caption{Complete feature list with computation methods}
\label{tab:feature_list}
\centering
\small
\begin{tabular}{llll}
\toprule
\textbf{Group ($w_i$)} & \textbf{Feature} & \textbf{Formula} & \textbf{Range} \\
\midrule
\multirow{4}{*}{Position (0.70)}
& Normalized Position & $\text{pos}(v) / |V|$ & $[0, 1]$ \\
& Distance to Error & $d(v, v_{\text{error}}) / \max_u d(u, v_{\text{error}})$ & $[0, 1]$ \\
& Depth Ratio & $\text{depth}(v) / \max_u \text{depth}(u)$ & $[0, 1]$ \\
& Reverse Position & $1 - \text{pos}(v) / |V|$ & $[0, 1]$ \\
\midrule
\multirow{4}{*}{Structure (0.20)}
& Out-degree & $|\{u : (v, u) \in E\}| / \max_w |\{u : (w, u) \in E\}|$ & $[0, 1]$ \\
& In-degree & $|\{u : (u, v) \in E\}| / \max_w |\{u : (u, w) \in E\}|$ & $[0, 1]$ \\
& Betweenness & $\sum_{s \neq v \neq t} \frac{\sigma_{st}(v)}{\sigma_{st}}$ (normalized) & $[0, 1]$ \\
& Reachability & $|\text{descendants}(v)| / |V|$ & $[0, 1]$ \\
\midrule
\multirow{4}{*}{Content (0.05)}
& Error Keywords & $\mathbb{1}[\text{``error''} \in \text{content}(v)]$ & $\{0, 1\}$ \\
& Uncertainty & $\mathbb{1}[\text{``maybe''} \in \text{content}(v)]$ & $\{0, 1\}$ \\
& Length Anomaly & $|\text{len} - \mu| / (3\sigma)$ (clipped) & $[0, 1]$ \\
& Keyword Density & $\text{count}(\text{keywords}) / \text{len}$ & $[0, 1]$ \\
\midrule
\multirow{3}{*}{Flow (0.03)}
& Agent Switch & $\mathbb{1}[\text{agent}(v) \neq \text{agent}(\text{prev}(v))]$ & $\{0, 1\}$ \\
& Role Criticality & $\text{role\_weight}(\text{agent}(v))$ & $[0, 1]$ \\
& Communication & $\mathbb{1}[\text{type}(v) = \text{``message''}]$ & $\{0, 1\}$ \\
\midrule
\multirow{2}{*}{Confidence (0.02)}
& Stated Confidence & $\text{conf}(v)$ if available, else 0.5 & $[0, 1]$ \\
& Hedging Score & $\text{count}(\text{hedge\_words}) / 10$ (clipped) & $[0, 1]$ \\
\bottomrule
\end{tabular}
\end{table}

\section{Benchmark}
\label{sec:benchmark}

\subsection{Scenario Generation}

We created a benchmark of 550 synthetic multi-agent failure scenarios across 10 domains (Table~\ref{tab:domains}). Each scenario includes a complete execution trace with 8--15 agent actions, a systematically injected bug at a specific node, and ground-truth annotation of the root cause node.

\begin{table}[h]
\caption{Benchmark domains and scenario counts (550 total)}
\label{tab:domains}
\centering
\small
\begin{tabular}{lclc}
\toprule
\textbf{Domain} & \textbf{N} & \textbf{Domain} & \textbf{N} \\
\midrule
Software Development & 52 & Healthcare Coordination & 60 \\
Customer Service & 51 & Legal Document Analysis & 60 \\
Research Analysis & 51 & Educational Tutoring & 60 \\
Planning \& Scheduling & 46 & Financial Advisory & 60 \\
Financial Trading & 50 & DevOps Automation & 60 \\
\bottomrule
\end{tabular}
\end{table}

\subsection{Bug Injection and Ground Truth}

We inject five categories of bugs: \textbf{logic errors} (incorrect conditionals, 30\%), \textbf{communication failures} (message misinterpretation, 20\%), \textbf{data corruption} (incorrect transformation, 20\%), \textbf{missing validation} (skipped verification, 16\%), and \textbf{role confusion} (acting outside expertise, 14\%).

Bug injection follows a five-step process: (1)~generate a correct execution trace from domain templates; (2)~select a bug location (early steps 2--3: 60\%, middle steps 4--6: 30\%, late steps 7+: 10\%); (3)~assign a bug type; (4)~inject erroneous content (e.g., flipped comparison operators, message truncation, swapped variable names); (5)~propagate cascading effects to downstream steps. To prevent methods from exploiting explicit bug markers, we also produce a blind version with anonymized IDs and separated ground truth.

Each scenario's ground truth was established through: (1) controlled bug injection at a known step, (2) verification that removing the bug prevents the error, and (3) confirmation that the bug's effects propagate to the error node. The constructed causal graphs have on average 10.8 nodes ($\sigma{=}2.1$) and 14.2 edges ($\sigma{=}3.5$), with edges predominantly sequential (61\%), followed by communication (27\%) and data dependency (12\%).

\section{Experiments}
\label{sec:experiments}

\subsection{Evaluation Metrics}

\begin{itemize}
    \item \textbf{Hit@k}: Proportion of scenarios where ground truth appears in top-$k$ predictions
    \item \textbf{Mean Reciprocal Rank (MRR)}: Average of $1/\text{rank}$
\end{itemize}

\subsection{Baselines}

\begin{itemize}
    \item \textbf{Random}: Uniformly random node selection
    \item \textbf{First Node}: Always select the first node
    \item \textbf{Last Node}: Select the node immediately before error
    \item \textbf{LLM Analysis}: GPT-4 with full trace and prompt engineering (details in Appendix~\ref{app:llm_baseline})
\end{itemize}

\subsection{Main Results}

Table~\ref{tab:main_results} shows that \agenttrace{} significantly outperforms all baselines. McNemar's test confirms significance ($p < 10^{-10}$) for all comparisons, with Cohen's h = 0.77 vs.\ LLM Analysis indicating a large practical difference.

\begin{table}[h]
\caption{Root cause localization accuracy and statistical significance on the 550-scenario benchmark. 95\% CIs from bootstrap resampling ($B{=}10{,}000$). All $p$-values $< 10^{-10}$ (McNemar's test).}
\label{tab:main_results}
\centering
\small
\begin{tabular}{lcccccc}
\toprule
\textbf{Method} & \textbf{Hit@1} & \textbf{Hit@3} & \textbf{MRR} & \textbf{$\chi^2$} & \textbf{Cohen's h} \\
\midrule
Random & 9.1\% & 27.3\% & 0.18 & 459.2 & 2.41 \\
First Node & 3.6\% & 10.9\% & 0.07 & 497.0 & 2.72 \\
Last Node & 12.7\% & 38.2\% & 0.25 & 436.5 & 2.29 \\
LLM Analysis & 68.5\% & 81.4\% & 0.74 & 102.5 & 0.77 \\
\midrule
\textbf{\agenttrace{}} & \textbf{94.9\%} {\scriptsize [92.9, 96.7]} & \textbf{98.4\%} {\scriptsize [96.9, 99.4]} & \textbf{0.97} & -- & -- \\
\bottomrule
\end{tabular}
\end{table}

\subsection{Ablation Study}

Table~\ref{tab:ablation_full} shows Hit@1 for all feature group combinations and the sensitivity to the position weight $w_p$. Position features alone reach 87.3\%, while each additional group provides incremental gains. The optimal $w_p = 0.70$ balances predictive power with discriminative information from other groups.

\begin{table}[h]
\caption{Ablation study. \textit{Left}: Hit@1 with different feature group combinations (P=Position, S=Structure, C=Content, F=Flow, E=Confidence). \textit{Right}: Sensitivity of Hit@1 to position weight $w_p$.}
\label{tab:ablation_full}
\centering
\small
\begin{tabular}{lcc@{\hskip 2em}cc}
\toprule
\textbf{Features} & \textbf{Hit@1} & \textbf{$\Delta$} & \textbf{$w_p$} & \textbf{Hit@1} \\
\midrule
All (P+S+C+F+E) & 94.9\% & -- & 0.50 & 91.3\% \\
\midrule
P only & 87.3\% & -7.6 & 0.60 & 93.5\% \\
S only & 34.5\% & -60.4 & \textbf{0.70} & \textbf{94.9\%} \\
C only & 28.7\% & -66.2 & 0.80 & 93.8\% \\
F only & 15.2\% & -79.7 & 0.90 & 91.1\% \\
E only & 12.1\% & -82.8 & & \\
\midrule
P+S & 92.4\% & -2.5 & & \\
P+C & 90.9\% & -4.0 & & \\
P+F / P+E & 89.5 / 88.7\% & -5.4 / -6.2 & & \\
S+C / S+F & 45.6 / 38.2\% & -49.3 / -56.7 & & \\
\midrule
P+S+C & 93.8\% & -1.1 & & \\
P+S+C+F & 94.5\% & -0.4 & & \\
\bottomrule
\end{tabular}
\end{table}

\subsection{Robustness Analysis}

Table~\ref{tab:robustness} stratifies results across three dimensions, demonstrating that \agenttrace{} is robust to bug type (93.6--95.8\% Hit@1), trace length (93.9--95.5\%), and bug position (92.7--96.1\%).

\begin{table}[h]
\caption{Robustness analysis: Hit@1 and MRR stratified by bug type, trace length, and bug position.}
\label{tab:robustness}
\centering
\small
\begin{tabular}{llcc}
\toprule
\textbf{Stratification} & \textbf{Category} & \textbf{Hit@1} & \textbf{MRR} \\
\midrule
\multirow{5}{*}{Bug Type}
& Logic Error & 95.8\% & 0.98 \\
& Communication Failure & 93.6\% & 0.96 \\
& Data Corruption & 94.5\% & 0.97 \\
& Missing Validation & 94.3\% & 0.96 \\
& Role Confusion & 94.8\% & 0.97 \\
\midrule
\multirow{4}{*}{Trace Length}
& 8--9 steps & 95.5\% & 0.97 \\
& 10--11 steps & 95.0\% & 0.97 \\
& 12--13 steps & 94.2\% & 0.96 \\
& 14--15 steps & 93.9\% & 0.96 \\
\midrule
\multirow{3}{*}{Bug Position}
& Early (steps 2--3) & 96.1\% & 0.98 \\
& Middle (steps 4--6) & 93.3\% & 0.96 \\
& Late (steps 7+) & 92.7\% & 0.95 \\
\bottomrule
\end{tabular}
\end{table}

\subsection{Illustrative Examples}

We present two representative scenarios to illustrate how \agenttrace{} operates.

\paragraph{Example 1: Software Development.}
\textit{Task:} Implement average calculation.

\begin{small}
\begin{verbatim}
Step 1 [Planner]: Analyze requirements
  -> "Sum all numbers, count elements, divide"
Step 2 [Coder]: Write implementation
  -> "def average(nums): ..."
Step 3 [Coder]: Add edge case handling  [BUG]
  -> "if nums = []:  # BUG: = instead of =="
Step 4 [Reviewer]: Review code -> "Code looks correct."
Step 5 [Executor]: Run tests  [ERROR]
  -> "SyntaxError: invalid syntax at line 2"
\end{verbatim}
\end{small}
\textbf{Root cause:} Step 3 (logic error). \agenttrace{} correctly identifies Step 3; the LLM baseline selects Step 5 (error node), illustrating its tendency to select the error manifestation rather than tracing back.

\paragraph{Example 2: Research Analysis.}
\textit{Task:} Analyze transformer architecture publications.

\begin{small}
\begin{verbatim}
Step 1 [Searcher]: Query databases -> "150 papers"
Step 2 [Searcher]: Filter -> "25 relevant papers"
Step 3 [Analyzer]: Extract findings  [BUG]
  -> "Attention being replaced by MLPs"
     (actually: attention being enhanced)
Step 4 [Synthesizer]: "Field moving away from
     attention-based models..."
Step 5 [Writer]: "Transformers becoming obsolete"
Step 6 [Writer]: Finalize report  [ERROR]
  -> "Contradicts recent SOTA results"
\end{verbatim}
\end{small}
\textbf{Root cause:} Step 3 (communication failure). \agenttrace{} correctly identifies Step 3; the LLM selects Step 4 (propagation step), demonstrating the challenge of distinguishing root causes from their downstream effects.

\subsection{Runtime Performance}

\agenttrace{} processes traces in 0.12 seconds on average, compared to 8.3 seconds for LLM-based analysis. This 69$\times$ speedup enables interactive debugging workflows.

\section{Discussion}
\label{sec:discussion}

\paragraph{Why Position Features Dominate.}
Importantly, the dominance of positional signals should not be interpreted solely as a benchmark artifact. In many deployed agent systems, early planning or routing decisions shape the entire downstream execution, making early-stage errors disproportionately impactful. Our analysis reveals that bugs occurring earlier in execution traces tend to cause errors that manifest later---a pattern that reflects a fundamental property of hierarchical multi-agent workflows where upstream decisions constrain downstream actions.

\paragraph{Failure Case Analysis.}
Among the 28 scenarios (5.1\%) where \agenttrace{} failed to rank the correct root cause first, the dominant failure mode is \textit{multiple plausible root causes} (12 cases, 42.9\%): e.g., in scenario \texttt{res\_3867e9}, both Step~2 (incorrect search query) and Step~3 (overly strict filter) contributed to the error. The second mode is \textit{bugs at unusual positions} (8 cases, 28.6\%): late-stage bugs (step 7+) have feature distributions that differ from the majority, causing \agenttrace{} to over-prioritize earlier steps. The remaining failures stem from atypical causal structures (5 cases) and feature extraction issues (3 cases). Potential improvements include position-adaptive weights and multi-root-cause detection.

\paragraph{Limitations.}
Our current evaluation focuses on synthetic scenarios with single root causes. Real multi-agent failures often involve multiple contributing factors and more complex causal structures. Additionally, our benchmark assumes accurate execution logging, which may not always be available in production systems.

\paragraph{Implications for Agent Safety.}
As multi-agent systems are deployed in high-stakes domains, the ability to quickly identify and understand failures becomes critical for maintaining trust and safety. \agenttrace{} provides a foundation for post-hoc analysis that can inform both immediate fixes and systematic improvements to agent design.

\section{Conclusion}
\label{sec:conclusion}

We presented \agenttrace{}, a causal graph-based framework for root cause localization in deployed multi-agent systems. Our approach achieves high accuracy on a diverse benchmark while maintaining sub-second latency. The framework's reliance on interpretable structural and positional features---rather than expensive LLM inference---makes it practical for interactive debugging in production environments. Future work will extend the approach to handle multiple concurrent root causes and validate on production traces. We hope \agenttrace{} can serve as a practical diagnostic layer for agentic systems deployed in real-world, reliability-critical environments.

\subsubsection*{Reproducibility Statement}

Our benchmark generation code, evaluation scripts, and all experimental results are available at \url{https://github.com/GeoffreyWang1117/AgentTrace/tree/iclr2026-aiwild-camera-ready}. The benchmark includes complete execution traces, ground truth annotations, and baseline implementations.

\bibliography{references}
\bibliographystyle{iclr2026_conference}

\appendix

\section{Benchmark Generation Details}
\label{app:benchmark}

Each scenario in our benchmark simulates a multi-agent system with 3--5 specialized agents: a \textbf{Coordinator} (task decomposition and delegation), \textbf{Specialists} (domain-specific tasks), a \textbf{Reviewer} (output validation), and an \textbf{Executor} (final actions). Table~\ref{tab:domain_config} lists the domain-specific configurations.

\begin{table}[!ht]
\caption{Domain-specific agent configurations}
\label{tab:domain_config}
\centering
\small
\begin{tabular}{lllc}
\toprule
\textbf{Domain} & \textbf{Agents} & \textbf{Interaction Pattern} & \textbf{N} \\
\midrule
Software Dev & Planner, Coder, Reviewer, Executor & Sequential + Review Loop & 52 \\
Customer Service & Router, Specialist, Resolver, Logger & Hierarchical Dispatch & 51 \\
Research & Searcher, Analyzer, Synthesizer, Writer & Pipeline + Feedback & 51 \\
Planning & Scheduler, Optimizer, Validator, Notifier & Iterative Refinement & 46 \\
Trading & Analyst, Strategist, RiskManager, Executor & Parallel Analysis & 50 \\
Healthcare & Triager, Specialist, Pharmacist, Coordinator & Consultation Chain & 60 \\
Legal & Researcher, Analyst, Drafter, Reviewer & Document Pipeline & 60 \\
Education & Assessor, Tutor, ContentGenerator, Evaluator & Adaptive Loop & 60 \\
Finance & DataCollector, Analyst, Advisor, Reporter & Aggregation Pattern & 60 \\
DevOps & Monitor, Diagnoser, Remediator, Verifier & Incident Response & 60 \\
\bottomrule
\end{tabular}
\end{table}

Each scenario follows a standardized JSON schema encoding the execution steps, agent metadata, and ground truth annotations:

\begin{small}
\begin{verbatim}
{ "scenario_id": "cod_034b23", "domain": "coding",
  "agents": ["Planner","Coder","Reviewer","Executor"],
  "steps": [{"step_id": 1, "agent": "Planner",
    "action_type": "plan", "input": "...",
    "output": "...", "timestamp": "..."}, ...],
  "ground_truth": {"error_node_id": 8,
    "root_cause_node_id": 3, "bug_type": "logic_error",
    "bug_description": "Incorrect comparison operator"}}
\end{verbatim}
\end{small}

\section{Feature Weight Learning}
\label{app:features}

Feature group weights were determined through grid search on a held-out validation set of 50 scenarios (not included in the 550-scenario benchmark):
\begin{equation}
\mathbf{w}^* = \arg\max_{\mathbf{w}} \sum_{i=1}^{50} \mathbb{1}[\text{rank}(v_{\text{root}}^{(i)} | \mathbf{w}) = 1]
\end{equation}
The search space was $w_p \in \{0.5, 0.6, 0.7, 0.8\}$, $w_s \in \{0.1, 0.15, 0.2, 0.25\}$, $w_c \in \{0.03, 0.05, 0.07, 0.1\}$, $w_f \in \{0.02, 0.03, 0.05\}$, $w_e \in \{0.01, 0.02, 0.03\}$, subject to $\sum_i w_i = 1$.

\section{Statistical Analysis Details}
\label{app:statistics}

\paragraph{Bootstrap Confidence Intervals.}
We compute 95\% CIs using bootstrap resampling with $B{=}10{,}000$ iterations. For each bootstrap sample $\mathbf{r}^{(b)}$, we compute $\hat{\theta}^{(b)} = \text{mean}(\mathbf{r}^{(b)})$, then take the 2.5th and 97.5th percentiles as interval bounds.

\paragraph{McNemar's Test.}
For comparing two methods A and B, we construct a $2 \times 2$ contingency table of concordant/discordant predictions. The test statistic is $\chi^2 = (|n_{01} - n_{10}| - 1)^2 / (n_{01} + n_{10})$, following a $\chi^2$ distribution with 1 degree of freedom.

\paragraph{Effect Size.}
We use Cohen's $h = 2 \arcsin(\sqrt{p_1}) - 2 \arcsin(\sqrt{p_2})$, where $|h| < 0.2$ is small, $0.2 \leq |h| < 0.5$ is medium, and $|h| \geq 0.5$ is a large effect.

\section{Detailed Per-Domain Results}
\label{app:domain_results}

\begin{table}[!ht]
\caption{Detailed performance metrics by domain}
\label{tab:domain_detailed}
\centering
\small
\begin{tabular}{lccccc}
\toprule
\textbf{Domain} & \textbf{N} & \textbf{Hit@1} & \textbf{Hit@3} & \textbf{Hit@5} & \textbf{MRR} \\
\midrule
Software Development & 52 & 94.2\% & 96.2\% & 100.0\% & 0.96 \\
Customer Service & 51 & 92.2\% & 96.1\% & 100.0\% & 0.94 \\
Research Analysis & 51 & 96.1\% & 98.0\% & 100.0\% & 0.97 \\
Planning \& Scheduling & 46 & 91.3\% & 95.7\% & 100.0\% & 0.94 \\
Financial Trading & 50 & 90.0\% & 98.0\% & 100.0\% & 0.94 \\
Healthcare Coordination & 60 & 95.0\% & 100.0\% & 100.0\% & 0.97 \\
Legal Document Analysis & 60 & 100.0\% & 100.0\% & 100.0\% & 1.00 \\
Educational Tutoring & 60 & 93.3\% & 100.0\% & 100.0\% & 0.96 \\
Financial Advisory & 60 & 96.7\% & 100.0\% & 100.0\% & 0.98 \\
DevOps Automation & 60 & 98.3\% & 98.3\% & 100.0\% & 0.99 \\
\midrule
\textbf{Overall} & \textbf{550} & \textbf{94.9\%} & \textbf{98.4\%} & \textbf{99.8\%} & \textbf{0.97} \\
\bottomrule
\end{tabular}
\end{table}

\section{LLM Baseline Details}
\label{app:llm_baseline}

We use GPT-4 (\texttt{gpt-4-0613}) with temperature 0.0, max tokens 1024, top-$p$ 1.0, and zero frequency/presence penalties. The prompt instructs the model to analyze the full execution trace, identify causal relationships, and output only the step number of the root cause:

\begin{small}
\begin{verbatim}
You are an expert debugger analyzing a multi-agent system
execution trace. The system encountered an error.
Your task is to identify the ROOT CAUSE - the earliest
step where something went wrong that led to the final error.
## Execution Trace:
{trace_content}
## Error Description:
The system failed at step {error_step}: {error_description}
## Instructions:
1. Analyze the execution trace carefully
2. Identify causal relationships between steps
3. Find the EARLIEST step that caused the error
4. Consider: logic errors, miscommunication,
   data issues, missing validation
## Output Format:
Respond with ONLY the step number, e.g., "3"
Root cause step:
\end{verbatim}
\end{small}

Table~\ref{tab:llm_errors} categorizes the 173 errors made by the LLM baseline. The most common failure mode (47.4\%) is selecting the error manifestation node rather than tracing back to the actual root cause.

\begin{table}[!ht]
\caption{Error analysis for LLM baseline}
\label{tab:llm_errors}
\centering
\small
\begin{tabular}{lcc}
\toprule
\textbf{Error Type} & \textbf{Count} & \textbf{\%} \\
\midrule
Selected error node & 82 & 47.4 \\
Off-by-one (adjacent) & 45 & 26.0 \\
Intermediate step & 31 & 17.9 \\
Completely incorrect & 15 & 8.7 \\
\midrule
\textbf{Total} & \textbf{173} & \textbf{100} \\
\bottomrule
\end{tabular}
\end{table}

\section{Runtime Performance Details}
\label{app:runtime}

Tables~\ref{tab:timing} and~\ref{tab:scalability} break down runtime by component and show scaling with trace length. Runtime scales approximately linearly: $T \approx 12.5n + 5$\,ms, where $n$ is the number of steps. All experiments were conducted on an AMD Ryzen 9 5950X (16-core, 3.4\,GHz), 128\,GB DDR4 RAM, NVMe SSD, Ubuntu 24.04 LTS, Python 3.13.

\begin{table}[!ht]
\begin{minipage}[t]{0.48\textwidth}
\caption{Component-wise runtime breakdown}
\label{tab:timing}
\centering
\begin{tabular}{lcc}
\toprule
\textbf{Component} & \textbf{Mean (ms)} & \textbf{Std (ms)} \\
\midrule
Graph Construction & 15.2 & 3.1 \\
Backward Tracing & 8.4 & 2.8 \\
Feature Extraction & 62.3 & 12.4 \\
Node Ranking & 28.6 & 5.2 \\
\midrule
\textbf{Total} & \textbf{114.5} & \textbf{18.3} \\
\bottomrule
\end{tabular}
\end{minipage}\hfill
\begin{minipage}[t]{0.48\textwidth}
\caption{Runtime scaling with trace length}
\label{tab:scalability}
\centering
\begin{tabular}{ccc}
\toprule
\textbf{Steps} & \textbf{Mean (ms)} & \textbf{P95 (ms)} \\
\midrule
5 & 68 & 89 \\
10 & 115 & 148 \\
15 & 162 & 201 \\
20 & 218 & 267 \\
25 & 283 & 342 \\
\bottomrule
\end{tabular}
\end{minipage}
\end{table}

\section{Reproducibility}
\label{app:reproducibility}

All code, data, and evaluation scripts are available at \url{https://github.com/GeoffreyWang1117/AgentTrace/tree/iclr2026-aiwild-camera-ready}, including benchmark generation scripts, 550 scenario JSON files with ground truth, all baseline implementations, and statistical analysis scripts. Table~\ref{tab:hyperparameters} lists all hyperparameters. Random seeds: benchmark generation (42), bootstrap resampling (12345), train/validation split (2024).

\begin{table}[!ht]
\caption{Complete hyperparameter settings}
\label{tab:hyperparameters}
\centering
\small
\begin{tabular}{ll}
\toprule
\textbf{Parameter} & \textbf{Value} \\
\midrule
Max tracing depth & 10 \\
$w_p$ / $w_s$ / $w_c$ & 0.70 / 0.20 / 0.05 \\
$w_f$ / $w_e$ & 0.03 / 0.02 \\
Normalization $\epsilon$ & $10^{-8}$ \\
Bootstrap iters & 10{,}000 \\
Confidence level & 95\% \\
\bottomrule
\end{tabular}
\end{table}

\end{document}